\documentclass[11pt]{article}
\usepackage[preprint]{acl}
\usepackage{times}
\usepackage{latexsym}
\usepackage[T1]{fontenc}
\usepackage[utf8]{inputenc}
\usepackage{microtype}
\usepackage{inconsolata}

\usepackage{graphicx}
\usepackage{xcolor}
\usepackage{amsmath,amssymb}
\newcommand{\Description}[1]{} 
\usepackage{titletoc} 
\AtBeginDocument{\renewcommand{\cite}{\citep}} 

\usepackage{pifont}
\usepackage{booktabs}
\usepackage{array}
\newcolumntype{L}[1]{>{\raggedright\arraybackslash}p{#1}} 
\usepackage{algorithm}
\usepackage{algorithmic}
\usepackage{xspace}
\usepackage{colortbl} 
\usepackage{soul}     
\usepackage{enumitem} 
\usepackage{tikz}
\usepackage{alltt}
\usepackage[most]{tcolorbox}
\newtcolorbox{codepanel}[3]{enhanced, colback=#1!70!white, colframe=#1!80!black, coltext=#2, title=#3, boxrule=0.6pt, arc=3pt, left=4pt, right=4pt, top=3pt, bottom=3pt, fonttitle=\bfseries\footnotesize, coltitle=#2,
  colbacktitle=#1!80!black, toptitle=1pt, bottomtitle=1pt}
\newcommand{\idb}[1]{{\setlength{\fboxsep}{0.6pt}\colorbox{black!16}{\strut #1}}}
\newcommand{\ido}[1]{{\setlength{\fboxsep}{0.6pt}\colorbox{black!28}{\textcolor{black}{\strut #1}}}}
\newcommand{\bs}{\textbackslash}
\definecolor{figblue}{HTML}{2A78D6}\definecolor{figorange}{HTML}{EB6834}

\newtcolorbox{rqsummary}[1]{enhanced, colback=gray!5!white,colframe=gray!5!black,reset,
boxsep=2pt,left=1pt,right=1pt,top=3pt,bottom=2pt,arc=0pt,boxrule=0pt,bottomrule=0.5pt,toprule=0.5pt}
\usetikzlibrary{positioning,arrows.meta,calc,shapes.geometric,fit}
\sethlcolor{yellow}

\definecolor{rowtint}{gray}{0.94}
\definecolor{headtint}{gray}{0.88}
\newcommand{\tabhead}[1]{\textbf{#1}}

\usepackage{soul}

\newcommand{\BfPara}[1]{{\noindent\ul{\bf#1.}}\xspace}
\newcommand*\cibbox[1]{\tikz[baseline=(char.base)]{\node[shape=rectangle,fill=gray!30!white,text=black,inner sep=1.5pt] (char) {#1};}}

\newcommand{\UlPara}[1]{\noindent\cibbox{{\bf{{#1.}}}}\xspace}
\newcommand*\cib[1]{\tikz[baseline=(char.base)]{ \node[shape=circle,fill=black,text=white,draw,inner sep=0.3pt] (char) {#1};}}
\setul{0.8pt}{}
\hypersetup{
	colorlinks=true,
	urlcolor=blue!70!black,
	linkcolor=blue!80!black,
	citecolor=blue!50!black,
}
\AtBeginDocument{%
}

\definecolor{headshade}{gray}{0.92}
\makeatletter
\newcommand{\shadedhead}[1]{%
  \@startsection{paragraph}{4}{\z@}%
    {0.8ex}%
    {-0.6em}%
    {\normalfont\normalsize}%
  *{\setlength{\fboxsep}{3pt}\colorbox{headshade}{\strut\textbf{#1}}}%
}
\makeatother
\definecolor{promptframe}{gray}{0.60}
\definecolor{promptback}{gray}{0.965}
\newsavebox{\promptboxbuf}
\newenvironment{promptbox}{%
  \par\vspace{0.5ex}\noindent
  \setlength{\fboxsep}{5pt}\setlength{\fboxrule}{0.5pt}%
  \begin{lrbox}{\promptboxbuf}%
  \begin{minipage}{\dimexpr\columnwidth-2\fboxsep-2\fboxrule\relax}%
  \footnotesize\setlength{\parindent}{0pt}%
  \setlength{\emergencystretch}{3em}\hbadness=10000 \raggedright}%
  {\end{minipage}%
  \end{lrbox}%
  \fcolorbox{promptframe}{promptback}{\usebox{\promptboxbuf}}%
  \par\vspace{0.7ex}}
\DeclareRobustCommand{\hlgray}[1]{{\setul{-1.6ex}{1.8ex}\setulcolor{gray!10}\ul{#1}}}

\newcommand{\ttc}[1]{{\texttt{\hlgray{#1}}}}
\newcommand{\tla}{{TLA$^+$}\xspace}

\title{ALIAS: Evaluating Performance and Memorization in LLM-Written \tla Specifications}

\author{\textbf{Arslan Bisharat, Mujtaba Nazari, Khushboo Bhadauria, Brian Ortiz,} \\
\textbf{Eric Spencer, Beatriz Santos, TaiNing Wang,} \\
\textbf{George K. Thiruvathukal, Konstantin L\"aufer, Mohammed Abuhamad} \\
Loyola University Chicago}

\begin{document}
\maketitle

\begin{abstract}
Memorization inflates the benchmark scores of large language models, since a model can reproduce answers it saw in training. Researchers test for it with consistent identifier renaming (CIR), which gives every identifier in a task a new name, and treat a lower score on the renamed task as evidence of recall. We test this method on the Temporal Logic of Actions Plus (\tla) language, whose model checker acts as an oracle by confirming whether a renamed task keeps the exact behaviors of the original. Across 111 \tla-Bench tasks, renaming lowers the correctness of Claude Opus 4.8 from 28\% to 18\% and of GPT-5 from 9\% to 7\%. Most of this decrease comes from how the test grades, which requires exact names that most descriptions never state. Grading by behavior shrinks the decrease to about 3 points for Opus 4.8 and to 0 for GPT-5. In two model families fine-tuned on disjoint halves of the tasks, renaming removes about half of the training gain in likelihood and costs 4.9 points (Qwen) and 1.3 points (Gemma) on unseen tasks. A lower score after renaming therefore mixes memorization, name-based grading and the cost of new names.
\end{abstract}

\section{Introduction} \label{sec:intro}
\tla is a formal specification language engineers use to find design errors in concurrent and distributed systems before they build them, and Amazon Web Services has reported subtle bugs found this way in DynamoDB and S3~\cite{newcombe2015aws}. A \tla specification describes the possible behaviors of a system, and the TLC model checker explores every reachable state and checks the properties that a configuration file lists by name~\cite{lamport2002specifying,tlcmodelcheck1999}. Writing a correct specification takes time and expertise, so LLMs that write specifications from natural-language descriptions could help more engineers use the method. Benchmarks measure this by checking each generated specification with the \tla parser SANY and the model checker TLC~\cite{sysmobench2026,formallm,tlabench2026}.

\tla-Bench and FormaLLM take their \tla specifications from public repositories on GitHub~\cite{formallm,tlabench2026}, so a model under test may have been trained on the specifications in these benchmarks, and a score may then measure recall as well as skill. Coding models memorize parts of their training data~\cite{yang2024memorization} and score higher on problems that resemble their training examples~\cite{riddell2024quantifying,matton2024leakage}. Public \tla data is scarce, since \tla-Bench collects 1,614 candidate specifications from 13 repositories~\cite{tlabench2026}, and it is unknown whether models recall these few specifications. Likelihood-based detectors perform near chance on training and held-out data from the same distribution~\cite{mink2024,duan2024mia}, so researchers rename every identifier of a task consistently and compare the scores on the original and renamed tasks~\cite{recode2023,whennames2025,codecocoon2026,putnamaxiom2025,gsmsymbolic2025,obfnng2026,exverus2026,tlaprover}. We call this \textbf{consistent identifier renaming (CIR)}, because each identifier gets one new name everywhere it appears. The test assumes that the renamed task is as hard as the original, so a lower score after CIR is read as evidence of memorization. We study this test on \tla, because TLC can confirm that a renamed task keeps the behaviors of the original and can grade an answer by behavior.

\begin{figure*}[t]
\centering
\resizebox{0.75\linewidth}{!}{
\begin{minipage}[t]{0.48\textwidth}
\begin{codepanel}{gray}{black}{Original description}
\footnotesize
This module specifies a simple reusable barrier synchronization protocol for N processes. Each process starts in control state \idb{"b0"} and may individually move to \idb{"b1"}; once all processes are in \idb{"b1"}, the barrier resets all processes back to \idb{"b0"} simultaneously. 
The specification defines the initial state, the next-state relation, a type correctness condition for the program counter function and a temporal barrier property stating that a process cannot leave the barrier while some other process has not yet entered it.

\vspace{3pt}\textbf{Configuration:}\\
\texttt{SPECIFICATION Spec}\\
\texttt{INVARIANT \idb{TypeOK}}\\
\texttt{PROPERTY \idb{BarrierProperty}}
\end{codepanel}
\end{minipage}~\hfill
\begin{minipage}[t]{0.48\textwidth}
\begin{codepanel}{gray!20!white}{black}{Renamed description}
\footnotesize
This module specifies a simple reusable barrier synchronization protocol for N processes. Each process starts in control state \ido{"Sb"} and may individually move to \ido{"Ir"}; once all processes are in \ido{"Ir"}, the barrier resets all processes back to \ido{"Sb"} simultaneously. The specification defines the initial state, the next-state relation, a type correctness condition for the program counter function and a temporal barrier property stating that a process cannot leave the barrier while some other process has not yet entered it.

\vspace{3pt}\textbf{Configuration:}\\
\texttt{SPECIFICATION Spec}\\
\texttt{INVARIANT \ido{Octvcq}}\\
\texttt{PROPERTY \ido{Bvwcdylhaazalut}}
\end{codepanel}
\end{minipage}
\Description{The original and renamed descriptions of the Barrier specification side by side, each with the configuration lines TLC checks. The descriptions differ only in the control-state labels. The configuration names TypeOK and BarrierProperty, renamed to Octvcq and Bvwcdylhaazalut, appear in neither description.}
}
\caption{The \ttc{Barrier} task as the model receives it, with the configuration that checks the output. Renamed parts are shaded, and the description states neither required name.}
\label{fig:barrier}
\end{figure*}

We investigate memorization in LLM-written \tla through three experiments on 111 \tla-Bench tasks whose renamed copies pass SANY and TLC. Experiment 1 grades the outputs of Claude Opus 4.8~\cite{opus48} and GPT-5~\cite{gpt5} the way \tla-Bench grades them, with the configuration file of the reference specification, which often requires names that the description never states (illustrated in \autoref{fig:barrier}). Experiment 2 fine-tunes two copies each of Qwen2.5-Coder-14B and Gemma 4 31B on disjoint halves of the tasks, so each task is seen by one copy and unseen by the other, and compares the renaming decrease on seen and unseen tasks. Experiment 3 grades the outputs of Experiment 1 again by behavior, which ignores names. Together, the experiments show that a lower score after renaming has three possible causes, \cib{1} memorization of the original task, \cib{2} a grading practice that requires names the model is never given and \cib{3} the extra difficulty of working with unfamiliar random names. This paper makes four contributions. \cib{1} We show that much of the decrease after CIR on \tla-Bench comes from the grading procedure. \cib{2} We separate memorization from the cost of CIR with controlled fine-tuning on disjoint halves of the tasks, so the training exposure of every task is known. \cib{3} We measure the decrease after CIR with a grade that checks behavior and ignores names. \cib{4} We give recommendations for contamination tests and benchmark builders, and we release ALIAS, with the renamed tasks, the exposure split, all model outputs and the behavior-based grader.\footnote{\url{https://github.com/LUC-AI4FM/ALIAS}}

\section{Background and Related Work}
\label{sec:problem}

\UlPara{Background}\label{sec:background} \tla is a language that describes how a system behaves over time~\cite{lamport2002specifying}. A specification lists the variables of the system, the starting states and the steps that may change the variables, together with the properties that the system must satisfy (see \autoref{fig:hourclock} for a small clock). SANY parses the module for \tla syntax correctness and checks that every name in a specification is defined. TLC then explores every state that a finite instance can reach and checks invariants, which must hold in every state, and temporal properties, which must hold over whole behaviors~\cite{tlcmodelcheck1999} (see \autoref{app:properties}). A configuration file gives TLC the constant values, the specification formula and the property names. A behavior is a sequence of states, and a specification $G$ refines a specification $R$ if every behavior of $G$ is also a behavior of $R$. TLC checks this by running $G$ against the formula of $R$ after mapping the variables of $R$ to those of $G$.

TLC finds operators and constants by the names in the configuration file, so the line \mbox{\ttc{INVARIANT TypeOK}} asks TLC to check the operator called \ttc{TypeOK}. A \tla-Bench \emph{task} has a natural-language description, which the model receives, and a configuration file, which checks the model's output. The human-written \tla module behind each task is its \emph{reference specification}, and the model never sees it~\cite{tlabench2026}. An output passes only if it uses the exact name of every constant and property in the configuration file.

\UlPara{Names in Specifications} Authors of \tla specifications often use domain concepts when selecting identifier names. Human readers understand formal specifications better with meaningful names~\cite{finney1998z}. For example, in Läufer's \ttc{Microwave} specification~\cite{laufer2022microwave} the names \ttc{door}, \ttc{radiation} and \ttc{DoorSafety} tie the natural language description to the formal state. The configuration file also lists names, so a check that uses it requires those exact names.

\UlPara{Related Work}\label{sec:related} \BfPara{LLMs for formal specification} Several benchmarks let a checker grade formal artifacts from LLMs~\cite{hou2024llm4se}. For \tla, SysMoBench grades models of real systems with SANY, TLC, trace validation and invariant checks~\cite{sysmobench2026}. Bisharat et al.\ use the comments of each specification as its task description~\cite{formallm} and \tla-Bench adds LLM-written descriptions to 403 model-checked and 897 parse-only specifications~\cite{tlabench2026}. Interface names raise \tla-Bench correctness up to 11-fold~\cite{tlabench2026}. Spencer et al.\ rename modules to limit contamination~\cite{tlaprover}. Dafny, Lean and Verus benchmarks grade with verifiers~\cite{dafnybench2024,verina2025,autoverus2025}, and some avoid contamination by design~\cite{leandojo2023,livefmbench2026}.

\BfPara{Measuring contamination} Some studies search training data for contamination~\cite{yang2024memorization,riddell2024quantifying,matton2024leakage}. Other studies control contamination by training models themselves. Magar and Schwartz pretrain BERT with chosen test sets in the corpus and separate memorization from its use on the task~\cite{magar2022contamination}, Jiang et al.\ pretrain GPT-2 models with test inputs or full test examples in the corpus~\cite{jiang2024investigating}, and Carlini et al.\ show that memorization grows with model size and with the number of repeats of an example~\cite{carlini2023quantifying}. Our exposure experiment follows this design with fine-tuning and adds renaming. Oren et al.\ test whether a model prefers a benchmark's canonical example order~\cite{oren2024proving}.

\BfPara{Perturbation and renaming} Perturbation studies change a task and keep its meaning. Code studies rename identifiers among other changes~\cite{recode2023,yang2022alert,evoeval2024,codefort2024}, and mathematical benchmarks change variables and constants~\cite{gsmsymbolic2025,putnamaxiom2025}. Wu et al.\ note that counterfactual task variants may simply be harder~\cite{wu2024reasoning}. Le et al.\ remove identifier names from code-understanding tasks and find lower scores even on execution tasks~\cite{whennames2025}. De Koning et al.\ relate the decrease of each renamed program-repair task to a model's likelihood of the original code~\cite{codecocoon2026}. Lean and Verus studies rename verified code~\cite{obfnng2026,exverus2026}. These studies observe decreases on models with unknown training data, so how much of a decrease comes from memory, from the grade or from the new names is unknown.
\section{Methods}
\label{sec:design}

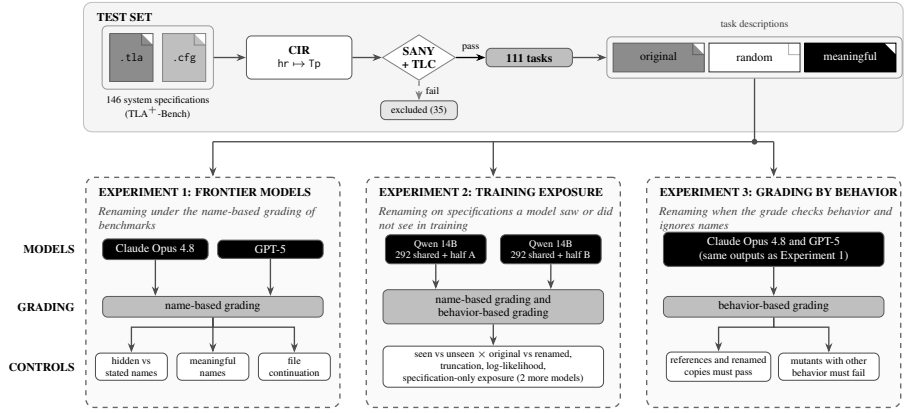
\begin{figure*}[t]
\centering
\resizebox{0.75\textwidth}{!}{
\begin{tikzpicture}[
  font=\scriptsize, >={Stealth[length=2mm]},
  box/.style={draw=black!65, fill=#1, rounded corners=3pt, align=center, inner sep=3pt, line width=0.6pt},
  model/.style={box=black, text=white},
  grade/.style={box=black!25},
  ctrl/.style={draw=black!70, fill=white, rounded corners=3pt, align=center, inner sep=2.5pt, line width=0.6pt},
  arr/.style={->, draw=black!70, line width=0.8pt},
  doc/.style={draw=#1!70!black, fill=#1, line width=0.6pt, minimum width=9mm, minimum height=10mm,
              align=center, inner sep=1pt, path picture={
                \fill[white] ([xshift=-2.4mm]path picture bounding box.north east) -- (path picture bounding box.north east)
                  -- ([yshift=-2.4mm]path picture bounding box.north east) -- cycle;
                \draw[#1!70!black, line width=0.6pt] ([xshift=-2.4mm]path picture bounding box.north east)
                  -- ([xshift=-2.4mm,yshift=-2.4mm]path picture bounding box.north east)
                  -- ([yshift=-2.4mm]path picture bounding box.north east);}},
  rq/.style={draw=black!60, dashed, rounded corners=5pt, fill=black!2, line width=0.8pt,
             minimum width=53mm, minimum height=48mm},
  rqtitle/.style={font=\bfseries\scriptsize, anchor=north west},
  q/.style={font=\itshape\scriptsize, text=black!75, align=left, anchor=north west, text width=50mm},
  ttl/.style={font=\bfseries\scriptsize}]

\begin{scope}[yshift=4.5mm]
\node[draw=black!25, fill=black!4, rounded corners=5pt, minimum width=171mm, minimum height=27mm] (T) at (85.5mm,-3.5mm) {};
\node[ttl, anchor=north west] at ($(T.north west)+(1.5mm,-1mm)$) {TEST SET};
\node[doc=black!45!white] (tla) at (10mm,-1.5mm) {\texttt{.tla}};
\node[doc=black!25!white] (cfg) at (21mm,-1.5mm) {\texttt{.cfg}};
\node[draw=black!40, rounded corners=3pt, inner sep=1.5mm, fit=(tla)(cfg)] (src) {};
\node[font=\tiny, align=center, anchor=north] at (src.south) {146 system specifications\\(\tla-Bench)};
\node[box=white, minimum width=22mm, minimum height=9mm] (ren) at (45mm,-1.5mm) {\textbf{CIR}\\\texttt{hr} $\mapsto$ \texttt{Tp}};
\node[draw=black!50, fill=white, diamond, aspect=1.6, inner sep=0.5pt, align=center, line width=0.6pt]
  (chk) at (70mm,-1.5mm) {\textbf{SANY}\\\textbf{+ TLC}};
\node[box=black!25, minimum width=18mm] (set) at (93mm,-1.5mm) {\textbf{111 tasks}};
\node[doc=black!45!white, minimum width=19mm, minimum height=6mm] (d1) at (120mm,-1.5mm) {original};
\node[doc=white, minimum width=19mm, minimum height=6mm, draw=black!80] (d2) at (140mm,-1.5mm) {random};
\node[doc=black, text=white, minimum width=19mm, minimum height=6mm] (d3) at (160mm,-1.5mm) {meaningful};
\node[font=\tiny, text=black!75, anchor=south] at (140mm,3mm) {task descriptions};
\node[font=\tiny, text=black!75, anchor=north] at (130mm,-6mm) {};
\node[font=\tiny, text=black!75, anchor=north] at (160mm,-6mm) {};
\draw[arr] (src.east) -- (ren.west);
\draw[arr] (ren.east) -- (chk.west);
\node[box=black!10, minimum width=16mm, font=\tiny] (ex) at (70mm,-12.5mm) {excluded (35)};
\draw[arr, draw=gray, dashed] (chk.south) -- node[right, text=black, font=\tiny]{fail} (ex.north);
\draw[arr, draw=black] (chk.east) -- node[above, text=black, font=\tiny]{pass} (set.west);
\node[draw=black!40, rounded corners=3pt, inner sep=1.2mm, fit=(d1)(d2)(d3)] (D) {};
\draw[arr] (set.east) -- (D.west);
\end{scope}

\node[font=\scriptsize\bfseries, anchor=east] at (-0.5mm,-37mm) {MODELS};
\node[font=\scriptsize\bfseries, anchor=east] at (-0.5mm,-49mm) {GRADING};
\node[font=\scriptsize\bfseries, anchor=east] at (-0.5mm,-61.5mm) {CONTROLS};

\node[rq] (R1) at (27mm,-46mm) {};
\node[rqtitle] at ($(R1.north west)+(1.5mm,-1mm)$) {EXPERIMENT 1: FRONTIER MODELS};
\node[q] at ($(R1.north west)+(1.5mm,-5mm)$) {Renaming under the name-based grading of benchmarks};
\node[model, minimum width=22mm] (opus) at (15mm,-37mm) {Claude Opus 4.8};
\node[model, minimum width=22mm] (gpt) at (39mm,-37mm) {GPT-5};
\node[grade, minimum width=46mm] (gn) at (27mm,-49mm) {name-based grading};
\node[ctrl, font=\tiny, minimum width=15mm] (c1a) at (10mm,-61.5mm) {hidden vs\\stated names};
\node[ctrl, font=\tiny, minimum width=15mm] (c1b) at (27mm,-61.5mm) {meaningful\\names};
\node[ctrl, font=\tiny, minimum width=15mm] (c1c) at (44mm,-61.5mm) {file\\continuation};
\draw[arr] (opus.south) -- (opus.south |- gn.north);
\draw[arr] (gpt.south) -- (gpt.south |- gn.north);
\draw[arr] (gn.south) -- ++(0,-2mm) -| (c1a.north);
\draw[arr] (gn.south) -- (c1b.north);
\draw[arr] (gn.south) -- ++(0,-2mm) -| (c1c.north);

\node[rq] (R2) at (85.5mm,-46mm) {};
\node[rqtitle] at ($(R2.north west)+(1.5mm,-1mm)$) {EXPERIMENT 2: TRAINING EXPOSURE};
\node[q] at ($(R2.north west)+(1.5mm,-5mm)$) {Renaming on specifications a model saw or did not see in training};
\node[model, font=\tiny, minimum width=21mm] (qA) at (73.5mm,-37mm) {Qwen 14B\\292 shared + half A};
\node[model, font=\tiny, minimum width=21mm] (qB) at (97.5mm,-37mm) {Qwen 14B\\292 shared + half B};
\node[grade, minimum width=46mm] (g2) at (85.5mm,-49mm) {name-based grading and\\behavior-based grading};
\node[ctrl, font=\tiny, minimum width=46mm] (h) at (85.5mm,-61.5mm)
  {seen vs unseen $\times$ original vs renamed,\\truncation, log-likelihood,\\specification-only exposure (2 more models)};
\draw[arr] (qA.south) -- (qA.south |- g2.north);
\draw[arr] (qB.south) -- (qB.south |- g2.north);
\draw[arr] (g2.south) -- (h.north);

\node[rq] (R3) at (144mm,-46mm) {};
\node[rqtitle] at ($(R3.north west)+(1.5mm,-1mm)$) {EXPERIMENT 3: GRADING BY BEHAVIOR};
\node[q] at ($(R3.north west)+(1.5mm,-5mm)$) {Renaming when the grade checks behavior and ignores names};
\node[model, minimum width=46mm] (m3) at (144mm,-37mm) {Claude Opus 4.8 and GPT-5\\(same outputs as Experiment 1)};
\node[grade, minimum width=46mm] (g3) at (144mm,-49mm) {behavior-based grading};
\node[ctrl, font=\tiny, minimum width=21mm] (c3a) at (132.5mm,-61.5mm) {references and renamed\\copies must pass};
\node[ctrl, font=\tiny, minimum width=21mm] (c3b) at (155.5mm,-61.5mm) {mutants with other\\behavior must fail};
\draw[arr] (m3.south) -- (g3.north);
\draw[arr] (g3.south) -- ++(0,-2mm) -| (c3a.north);
\draw[arr] (g3.south) -- ++(0,-2mm) -| (c3b.north);

\coordinate (bus) at ($(D.south |- T.south)+(0,-2mm)$);
\draw[draw=black!70, line width=0.8pt] (D.south) -- (bus);
\draw[draw=black!70, line width=0.8pt] (bus -| R1.north) -- (bus -| R3.north);
\draw[arr] (bus -| R1.north) -- (R1.north);
\draw[arr] (bus -| R2.north) -- (R2.north);
\draw[arr] (bus -| R3.north) -- (R3.north);
\fill[black!70] (bus) circle (0.6mm);
\end{tikzpicture}}
\Description{A test-set strip shows consistent identifier renaming of 146 TLA+-Bench system specifications and re-verification with SANY and TLC, which keeps 111 tasks with original and random-name descriptions and 77 with meaningful-name descriptions. Below it, three equal panels show Experiment 1 (frontier models graded by name, with three controls), Experiment 2 (two fine-tuned models on disjoint halves, graded both ways) and Experiment 3 (the same frontier outputs graded by behavior, with a grader validated on references, renamed copies and mutants).}
\caption{The test set (top) and the three experiments (bottom), each with its models, grade and controls.}
\label{fig:pipeline}
\end{figure*}

A CIR test renames the identifiers of each reference specification, rewrites the description and configuration file with the new names and compares a model's correctness on the two versions of each task, which we call the \emph{original} and \emph{renamed} conditions. We build such a test set and run three experiments on it with the same two grades (illustrated in \autoref{fig:pipeline}).

\UlPara{Dataset} \label{sec:design-testset} We use the human-written reference specifications of the model-checked tier of \tla-Bench~\cite{tlabench2026}. This tier contains 403 specifications, each with a configuration file that passes TLC, and labels 146 of them as system specifications. We exclude the other 257 utility modules because they are small helpers in which our renaming affects a median of only five identifiers. Our test set contains the 111 tasks whose renamed reference specifications pass both SANY and TLC (see \autoref{tab:counts}).

\BfPara{Renaming}
\label{sec:design-rename} We replace each author-defined module name, constant, variable and operator with a random name of the same length, because SANY reads the layout of a formula (see \autoref{app:renaming}). The same map renames the specification, its configuration file and every module that imports it, and SANY and TLC confirm that the renamed copy still parses and passes model checking.

\BfPara{Renamed descriptions} Each specification in \tla-Bench comes with a natural language description that mentions many of its operators and constants by name~\cite{tlabench2026}. The original condition gives the LLM this description, and the renamed condition prompts the LLM with a copy of the original description in which those identifiers are replaced. Replacing every renamed identifier in the text would also change unrelated words, because authors use common words and algorithm names as identifiers (see \autoref{app:renaming} for the example and further details). We therefore rename each description with a fixed rule of three parts. \cib{1} A name that the configuration refers to is always renamed, because the model needs the new name for TLC to check its output. \cib{2} A lowercase word or a \tla module name keeps its name. \cib{3} Every other identifier is renamed. This rule avoids damage to 45 descriptions.

\BfPara{Identifier kinds}\label{sec:problem-description} We sort the identifiers of the reference specifications of the 111 tasks into three categories, \cib{1} domain words, \cib{2} \tla conventions such as \ttc{Init} and \ttc{TypeOK} and \cib{3} abbreviations or single letter identifiers such as \ttc{hr} (see \autoref{app:idkinds}). \autoref{tab:idkinds} shows that a third of the 1,788 identifiers that our CIR test modifies are domain words. The model never sees the reference specification and reads only its English description. Renaming changes the identifiers that this description mentions by name and leaves the rest of the text as it is, so ordinary words such as \emph{barrier} or \emph{process} remain (illustrated in \autoref{fig:barrier}). Of the domain identifiers whose words the original description uses, 199/219 (91\%) keep those words after CIR.

\BfPara{Hidden required names}\label{sec:problem-grader} The configuration file of a task names \cib{1} the constants that TLC assigns values to, \cib{2} the invariants and properties that TLC checks and \cib{3} the constraints that limit the states TLC explores. An output is checked with this file, so it must define every identifier the file names. We call a listed identifier \emph{hidden} if the description given to the model never mentions it by name. In the \ttc{Barrier} task, the configuration checks \ttc{TypeOK} and \mbox{\ttc{BarrierProperty}}, and the description mentions neither of these two names (illustrated in \autoref{fig:barrier}). After renaming, the configuration asks for \ttc{Octvcq} and \ttc{Bvwcdylhaazalut}, so a correct output fails whether or not the model has seen the specification in training. We exclude \ttc{Init}, \ttc{Next} and \ttc{Spec}, because the prompt asks for them by name and the name-based grade accepts them under any name (see \autoref{app:grading}). In 78/111 tasks, the configuration file lists at least one hidden name, so a correct output must define that exact name to pass TLC, and 29 of these tasks list five or more. The 333 hidden names are 115 constants, 113 invariants, 87 properties and 18 constraints. With the original description, a model can still guess some hidden names without recall, because 77 follow \tla conventions such as \ttc{TypeOK}. After renaming, every hidden name is a random string that no description, \tla convention or published file contains. Renaming therefore changes two parts of a task, \cib{1} the description, which loses the author's identifiers but keeps most of its domain words, and \cib{2} the configuration file, which requires exact names.

\UlPara{Grading}\label{sec:design-grading} 

\BfPara{Parsing} A module passes parsing in our grading only if SANY exits with code 0 and reports no error. However, SANY can complete parsing with an exit code of 0 even for semantic errors such as an undefined operator.

\BfPara{Name-based grade} This grade follows \tla-Bench. An output is correct if it parses and passes TLC with the configuration file of its task, and a TLC run over 300 seconds counts as a failure. For a renamed task, each name in the configuration file is replaced by its new name from the renamed specification and description (see \autoref{app:grading}).

\BfPara{Behavior-based grade} This grade compares the behaviors of an output $G$ with those of the reference $R$ and does not require the names of the configuration file. It searches for a one-to-one mapping from the variables and constants of $R$ to those of $G$, with identical names first and then up to 60 permutations. Under a mapping, the forward check runs $G$ and checks the specification formula of $R$ and the invariants and properties of its configuration, with the definitions of $R$, so it passes when every behavior of $G$ is a behavior of $R$ and satisfies the properties of $R$. The reverse check runs $R$ and checks the safety formula $\mathit{Init} \land \Box[\mathit{Next}]_v$ of $G$ under the inverse mapping, so an output that allows too few behaviors, such as one that never moves, fails. $G$ is correct when both checks pass under one mapping (see \autoref{app:grading}). Variables whose string values act as control labels, such as the \ttc{pc} of a PlusCal translation, get extra handling (see \autoref{app:labels}). Each condition is graded against its own reference specification, so the renamed condition uses the renamed specification, its renamed imported modules and the renamed configuration.

\UlPara{Analysis and Metrics}\label{sec:design-analysis} 

\BfPara{Metrics} We report three metrics for each sample. \cib{1} \emph{Parse}, the output passes SANY with no syntax or semantic error. \cib{2} \emph{Correct, name-based}, the output parses and passes TLC under the configuration of its condition. \cib{3} \emph{Correct, behavior-based}, the output passes both refinement checks against the reference. We also report text similarity to the reference (see \autoref{app:similarity}). The unit of analysis is the task. For each task and condition, the pass rate is the fraction of the five samples that pass, which estimates pass@1~\cite{chen2021codex}. We report the mean difference between two conditions with a 95\% confidence interval from 10,000 bootstrap resamples of tasks. We also report a two-sided sign test over the tasks whose pass rates differ. For the two renamings, we test equivalence with two one-sided tests at a margin of 5 points. Equivalence holds when the 90\% bootstrap interval of the difference lies inside the margin.

\UlPara{Experiment 1: Name-Based Grading}\label{sec:exp1} This experiment measures how renaming changes the correctness of frontier models under the grading that \tla-Bench uses, and which tasks lose correctness.

\BfPara{Models and prompting} Renaming can only lower a score if the model writes some correct specifications for the original tasks. On \tla-Bench, open-weight models pass TLC on at most 1\% of system specifications and Claude Opus 4.5 passes on 16\%~\cite{tlabench2026}, so we study Claude Opus 4.8 and GPT-5. Both models receive the same prompt and write five independent samples per task and condition, and Experiment 3 grades the same outputs a second time. The prompt asks for a complete \tla module with \ttc{Init}, \ttc{Next}, \ttc{Spec} and the safety, liveness and fairness conditions of the description (see \autoref{app:prompt}).

\BfPara{Controls} Three controls test other explanations of a decrease. \cib{1} We split the results by whether a task hides a required name (illustrated in \autoref{fig:barrier}), because a hidden name fails a correct output in the renamed condition. \cib{2} Random names may lower correctness because they carry no meaning, so a third condition replaces each identifier with a meaningful name that Opus 4.8 proposes, re-verified with SANY and TLC in the same way, which gives 77 valid tasks (see \autoref{app:meaningful}). \cib{3} A file continuation test shows each model the first 40\% of a published file with its repository and path and asks it to continue the file, and we compare the longest token run shared with the real file against the same run for the renamed copy, which no model can have seen (see \autoref{app:continuation}).

\UlPara{Experiment 2: Controlled Exposure}\label{sec:exp2} Experiment 1 cannot tell whether a model saw a task, because the training data of Opus 4.8 and GPT-5 is unknown. The pretraining data of open models such as Qwen and Gemma is not fully known either, so this experiment adds exposure that we control. It fine-tunes each model on half of the tasks and compares the renaming decrease on seen and unseen tasks.

\BfPara{Fine-tuned models} We split the 111 tasks at random with seed 20260926 into halves, A (55 tasks) and B (56 tasks). Both training sets contain the 292 model-checked \tla-Bench tasks outside the test set. We fine-tune one copy of Qwen2.5-Coder-14B-Instruct on these 292 tasks and the 55 of half A, and a second copy on the 292 and the 56 of half B, and we shuffle each training set with a fixed seed. Each training pair consists of a declarative description and its reference specification. Every test task is thus seen by one model and unseen by the other. We call this task-pair exposure since the model sees each specification together with its test description. Both LLMs are fine-tuned with the Low-Rank Adaptation (LoRA) method~\cite{hu2022lora} and train for three epochs over their respective datasets, so each specification is seen three times in training (see \autoref{app:finetune} for the training settings). We test both models on the 110 tasks with a configuration in both conditions. We repeat the experiment with a second model family, Gemma 4 31B~\cite{gemma4}, fine-tuned on the same halves with the same training settings (see \autoref{app:gemma}). Each model generates five samples per task and condition with temperature 0.7, top-$p$ 0.95 and at most 1,024 new tokens per sample.

\BfPara{Randomized exposure} Crossing the two models with the two conditions gives four cells per task. From these cells we measure \cib{1} the exposure effect, the gain from training on a task, \cib{2} how much renaming reduces that gain and \cib{3} the renaming cost on unseen tasks.

\BfPara{Controls} Random names have more tokens, so renamed outputs more often hit the 1,024-token limit, and we repeat the analysis on the tasks with no truncated output (see \autoref{app:truncation}). A second pair of models sees each half of the test set as the raw \ttc{.tla} file alone (see \autoref{app:speconly}).

\BfPara{Log-likelihood} We also measure how likely each model finds the reference specification, given its description. This is the mean token log-likelihood, the average log-probability that the model assigns to each token of the reference, and a higher value means that the model expects that exact text more. The measure carries information even when few outputs pass TLC. We score the first 2,048 tokens of description and reference together, the length limit used in training. A gain in likelihood after exposure could come from the original names alone, so we also report the mean over the tokens that do not overlap a renamed identifier, which hold the keywords, operators and structure.

\BfPara{Analysis} The exposure effect in each condition is the mean over tasks of the pass rate of the model that saw the task minus the pass rate of the model that did not. The change in the exposure effect under renaming is the difference in differences (DiD)
\[
\mathrm{DiD} = \frac{1}{n}\sum_{s}\Big[(c_s - r_s)_{\text{seen}} - (c_s - r_s)_{\text{unseen}}\Big],
\]
where $c_s$ and $r_s$ are the pass rates (or mean log-likelihoods) of task $s$ in the original and renamed conditions, respectively. The first term uses the model that saw $s$ and the second uses the model that did not. A positive DiD means that renaming costs more on seen tasks. The renaming cost on unseen tasks is $\frac{1}{n}\sum_s (c_s - r_s)_{\text{unseen}}$, and memory plays no part in it.

\UlPara{Experiment 3: Behavior-Based Grading}\label{sec:exp3} Experiment 1 grades outputs by name, so a correct output fails whenever it misses a hidden name. This experiment grades the same outputs of Opus 4.8 and GPT-5 again by behavior and compares the decrease under both grades on the same tasks. A decrease that remains under the behavior-based grade cannot come from hidden names.

\BfPara{Grader validation}\label{sec:design-validation} As positive controls, the grade should accept each reference against itself, each renamed reference with its renamed imported modules and a copy of each reference with renamed operators that still passes SANY and TLC. As negative controls, it should reject mutants, which change one number, comparison, initial value, guard or disjunct of a reference, and we keep a mutant only if TLC reports a different number of reachable states (see \autoref{app:validation}). The grade covers the 90/111 tasks whose reference passes against itself, and both grades are compared on these tasks.

\BfPara{Disagreements} We read the outputs on which the two grades disagree and sort them by cause. As an upper bound on errors of our grader, we also recompute the decrease with every output that passes only the name-based grade counted as correct (see \autoref{app:disagree} for the cases).

\section{Results and Discussion}
\label{sec:results}

\UlPara{Experiment 1: Name-Based Grading}\label{sec:exp1-findings} \BfPara{Opus 4.8} Opus 4.8 is correct on 28.3\% of samples with the original descriptions and on 18.2\% with the renamed ones (\autoref{tab:overview}), a mean decrease per task of 10.1 points (95\% CI 5.0 to 15.7), with 18 tasks that now fail and two that now pass (sign test $p=0.0004$). The parse rate stays near 81\% in both conditions. In the original condition it would be 93.3\% if outputs that SANY accepts with exit code 0 despite a semantic error counted as parsed.

\begin{table}[t]
\centering
\footnotesize
\setlength{\tabcolsep}{4pt}
\resizebox{0.4\textwidth}{!}{
\begin{tabular}{lrrr}
\toprule
\tabhead{Measure} & \tabhead{Original} & \tabhead{Random} & \tabhead{Meaningful} \\
\midrule
\multicolumn{4}{l}{\emph{Claude Opus 4.8}} \\
Parse (SANY) & 80.4 & 81.6 & 81.0 \\
Correct, name & 28.3 & 18.2 & 20.0 \\
Correct, behavior & 38.4 & 35.3 & 37.7 \\
\midrule
\multicolumn{4}{l}{\emph{GPT-5}} \\
Parse (SANY) & 66.1 & 70.5 & 69.1 \\
Correct, name & 9.2 & 7.0 & 7.3 \\
Correct, behavior & 10.9 & 10.9 & 11.1 \\
\bottomrule
\end{tabular}}
\caption{Frontier models on the test set, in percent. Meaningful names use the 77 tasks with a valid meaningful renaming, and the behavior-based rows use the 90 that the grade covers.}
\label{tab:overview}
\end{table}

\BfPara{Hidden names} On the 78 tasks with a hidden required name, correctness is 11.3\% with the original descriptions and 0.0\% with the renamed ones, so all 14 tasks that passed now fail (see \autoref{fig:hidden}). On the 33 tasks whose descriptions state every required name, correctness is 68.5\% and 61.2\%, a decrease of 7.3 points (95\% CI $-2.4$ to 18.2), with four tasks that now fail and two that now pass.

\BfPara{Meaningful names} Under the name-based grade, meaningful and random names score the same in correctness within a margin of 5 points (see \autoref{app:meaningful}).

\BfPara{GPT-5} GPT-5 is correct on 9.2\% of samples with the original descriptions and 7.0\% with the renamed ones, a decrease of 2.2 points (95\% CI $-0.4$ to 4.7), and on the 78 tasks with a hidden name it falls from 2.3\% to 0.0\%.

\BfPara{File continuation} How much of a published file a model can continue does not predict which tasks lose correctness under renaming (see \autoref{app:continuation}).

\BfPara{What Experiment 1 shows} Renaming lowers the correctness of Opus 4.8 by about a third, which a CIR test reads as memorization. The decrease sits on the 78 tasks with hidden names, where the renamed configuration asks for random names that no model can know, and the 33 tasks that state every name lose 7.3 points with an interval that includes 0. Meaningful names and file continuation do not explain the decrease. Under the grading of \tla-Bench, a renaming decrease thus measures the grade as well as the model, and Experiment 3 tests how much.

\UlPara{Experiment 2: Controlled Exposure}\label{sec:exp2-findings} \BfPara{Qwen} Under the behavior-based grade, training on a task raises the correctness of Qwen by 7.6 points (95\% CI 3.8 to 12.0) with the original descriptions and by 4.2 points (95\% CI 1.1 to 7.8) with the renamed ones (\autoref{tab:hypotheses}; see also \autoref{fig:exposure}). Renaming lowers correctness by 4.9 points (95\% CI 2.0 to 8.4) on unseen tasks. The DiD is 3.3 points (95\% CI $-0.2$ to 7.3), and 3.5 points (95\% CI 0.4 to 6.9) under the name-based grade on 110 tasks.

\begin{table}[t]
\centering
\footnotesize
\setlength{\tabcolsep}{4pt}
\newcommand{\ci}[2]{\begin{tabular}[t]{@{}c@{}}#1\\[-1pt]{\scriptsize[#2]}\end{tabular}}
\resizebox{0.4\textwidth}{!}{
\begin{tabular}{lccc}
\toprule
\tabhead{Measure} & \tabhead{Exposure} & \tabhead{DiD} & \tabhead{Unseen cost} \\
\midrule
\multicolumn{4}{l}{\emph{Qwen2.5-Coder-14B}} \\
Correct, behavior & \ci{7.6}{3.8, 12.0} & \ci{3.3}{$-$0.2, 7.3} & \ci{4.9}{2.0, 8.4} \\
Correct, name & \ci{5.6}{2.2, 9.6} & \ci{3.5}{0.4, 6.9} & \ci{4.4}{1.5, 7.6} \\
Likelihood & \ci{0.34}{0.30, 0.39} & \ci{0.16}{0.13, 0.19} & -- \\
Likelihood, other & \ci{0.31}{0.27, 0.35} & \ci{0.07}{0.06, 0.09} & -- \\
\midrule
\multicolumn{4}{l}{\emph{Gemma 4 31B}} \\
Correct, behavior & \ci{8.0}{3.6, 12.9} & \ci{3.6}{$-$0.9, 8.2} & \ci{1.3}{$-$2.0, 4.9} \\
Correct, name & \ci{7.8}{3.6, 12.4} & \ci{5.8}{1.5, 10.5} & \ci{4.2}{0.4, 8.5} \\
Likelihood & \ci{0.33}{0.29, 0.38} & \ci{0.15}{0.12, 0.18} & -- \\
\bottomrule
\end{tabular}}
\caption{Effects of task-pair exposure for two model families with 95\% CIs, in percentage points for correctness and in mean token log-likelihood for likelihood. Unseen cost is original minus renamed on unseen tasks, and \emph{other} uses only tokens outside renamed identifiers.}
\label{tab:hypotheses}
\end{table}

\BfPara{Likelihood} For 108/110 tasks, the Qwen model trained on a task finds its reference specification more likely than the other model. Renaming lowers this exposure effect from 0.34 to 0.18, a DiD of 0.16 (95\% CI 0.13 to 0.19) that is positive for 110/110 tasks. On the tokens outside the renamed identifiers, the DiD is 0.07 (95\% CI 0.06 to 0.09) and positive for 101/110 tasks.

\BfPara{Gemma 4} Gemma 4 31B shows an exposure effect of 8.0 points (95\% CI 3.6 to 12.9) and a likelihood DiD of 0.15 (95\% CI 0.12 to 0.18), with a higher likelihood for the seen model on 109/110 tasks. Its renaming cost on unseen tasks is 1.3 points (95\% CI $-2.0$ to 4.9) under the behavior-based grade and 4.2 points (95\% CI 0.4 to 8.5) under the name-based grade (see \autoref{app:gemma}).

\BfPara{Controls} On the 43/90 tasks with no truncated output, the renaming cost on unseen tasks is 9.8 points (95\% CI 3.7 to 16.3; see \autoref{app:truncation}). Models fine-tuned on the raw \tla files show no effect on correctness, and their likelihood DiD is 0.08 (95\% CI 0.07 to 0.10; see \autoref{app:speconly}).

\BfPara{What Experiment 2 shows} With exposure known, the two parts of a decrease separate. Training leaves a memory of the seen tasks, and renaming removes about half of the likelihood gain in both families, including on keywords and structure that renaming leaves unchanged. On correctness, the DiD interval includes 0 under the behavior-based grade in both families, so the part of a correctness decrease that comes from memory is too small to measure on 90 tasks. Renaming also lowers correctness on tasks a model never saw, by 4.9 points for Qwen, where memory plays no part. A renaming decrease can therefore appear without memorization, and its size depends on the model.

\UlPara{Experiment 3: Behavior-Based Grading}\label{sec:exp3-findings} \BfPara{Grader validation} The grader accepts all 90 renamed references and all 56 valid operator-renamed copies, and rejects all 105 mutants with altered behavior (see \autoref{tab:validation}).

\BfPara{Opus 4.8} On the 90 tasks that both grades cover (see \autoref{fig:graders}), Opus 4.8 is correct on 31.8\% of samples with the original descriptions and on 21.3\% with the renamed ones under the name-based grade, a decrease of 10.4 points (95\% CI 4.7 to 16.9). Under the behavior-based grade, it is correct on 38.4\% and 35.3\%, a decrease of 3.1 points (95\% CI $-0.9$ to 7.3; see \autoref{app:exp3}). In the \ttc{Barrier} task, all five renamed outputs fail the name-based grade on the hidden names and pass the behavior-based grade.

\BfPara{Disagreements} Most outputs that pass only the behavior-based grade miss a hidden name. Outputs that pass only the name-based grade behave differently from the reference or hit a limit of our grader, and the different ones pass because TLC checks each invariant with the output's own definition, so even an output that defines \ttc{TypeOK} as \ttc{TRUE} passes that check (see \autoref{app:disagree}). If every name-only pass counts as correct, the decrease is 5.1 points (95\% CI 0.4 to 10.2) for Opus 4.8 and 0.4 points (95\% CI $-3.1$ to 4.0) for GPT-5.

\BfPara{GPT-5} On the same 90 tasks, the decrease of GPT-5 is 2.4 points under the name-based grade and 0.0 points (95\% CI $-3.6$ to 3.3) under the behavior-based grade, with 10.9\% in both conditions. Its 90\% interval, $-2.9$ to 2.9, lies inside the 5-point margin, but at a base rate near 11\% this margin is almost half of GPT-5's correctness.

\BfPara{What Experiment 3 shows} When the grade ignores names, the decrease of Opus 4.8 falls from 10.4 to 3.1 points and that of GPT-5 from 2.4 to 0.0 points, and for Opus 4.8 the 64 covered tasks with hidden names lose only 2.5 points (see \autoref{app:exp3}). Most of the decrease in Experiment 1 thus came from the grade. The interval for Opus 4.8 still reaches 7.3 points, so a cost of new names of the size that Experiment 2 measures on unseen tasks is still possible for Opus 4.8.

\UlPara{Discussion}\label{sec:discussion} \BfPara{Implications for {\tla} benchmarks} Experiments 1 and 3 show that a check with the reference configuration file grades both what a specification does and how it names identifiers, and in 78/111 of our tasks the description leaves out a required name. Listing every required name in the task separates the two, but it also tells the model which invariants and properties to write, which removes part of the task. The same check accepts outputs with weaker invariants (Experiment 3), and the SANY exit code overstates parsing (Experiment 1). Of four public \tla benchmarks, only \tla-Bench needs a name that the task does not state (see \autoref{tab:benchmarks}).

\BfPara{Implications for benchmark builders} From these results, we suggest four practices for \tla benchmarks, \cib{1} grade by behavior against the reference's own properties, \cib{2} when grading depends on names, state every required name in the task or report tasks with hidden names separately, \cib{3} count semantic errors as parse failures and \cib{4} re-verify renamed copies with the model checker. The same may hold outside \tla, because names carry meaning in code as well~\cite{whennames2025}, and unit tests and proof checkers also look up names that tasks may omit.

\BfPara{Implications for contamination studies} Experiment 2 shows that, under a grade that ignores names, a decrease after renaming can still come from two sources, memory of the original task and the cost of working with new names, and that known exposure separates them. Memory shows in the models we fine-tuned, since renaming removed about half of the gain from training in how likely the model finds the reference. The cost of new names shows on tasks a model never saw, since memory plays no part there, and under the behavior-based grade it was 4.9 points for Qwen and 1.3 points for Gemma (95\% CI $-2.0$ to 4.9). On a model whose training data is unknown, a single decrease adds these two sources together and cannot separate them. For Opus 4.8 and GPT-5, most of the decrease comes from grading, and the interval of the decrease under the behavior-based grade includes 0 but leaves room for a name cost of several points. Contamination studies could add a model with known exposure and report the decrease separately for seen and unseen items.

\section{Conclusion}
\label{sec:conclusion}

We renamed every author-defined identifier in 111 model-checked \tla specifications and confirmed with SANY and TLC that each renamed benchmark task keeps the behaviors of the original task. Under the grading of \tla-Bench, renaming lowers the correctness of Claude Opus 4.8 from 28.3\% to 18.2\%, mostly because of configuration names that the description never states. Graded by behavior, the decrease is 3.1 points for Opus 4.8 (95\% CI $-0.9$ to 7.3) and 0.0 points for GPT-5 (95\% CI $-3.6$ to 3.3). In the LLMs we fine-tuned, training made the reference specification more likely to the model, and renaming removed about half of that gain. On tasks a model never saw, renaming lowered correctness by 4.9 points for Qwen and 1.3 points for Gemma. A decrease after renaming therefore mixes memorization with grading by name and with the cost of new names, so it cannot show memorization by itself.

\section*{Limitations}

\BfPara{Internal validity} The descriptions shape what a model writes, and \tla-Bench writes them with an LLM. Every comparison in our experiments is paired, because the original and renamed versions of a task share one description and differ only in the renamed identifiers, so the quality of a description affects both conditions equally and does not change the decrease. The renaming rule keeps 91\% of the domain words of a description, so the renamed task asks for the same system. The rule also keeps the names of known algorithms such as Paxos, which a model could use to recall a known specification. Tasks with such a name are harder for both frontier models and lose no more correctness under renaming than the other tasks (see \autoref{app:algonames}), so this risk is small, and a direct test that asks for a specification from the algorithm name alone needs further study. Claude Opus 4.8 proposed the meaningful names and is also tested on them, so these names may suit it. In the exposure experiment, both models of a family share the same 292 training tasks with original names, so even unseen tasks follow naming conventions that the models learned.

\BfPara{Construct validity} We measure three constructs. \cib{1} \emph{Correctness} is passing TLC on the finite instance that the configuration bounds, so an output that passes may still be wrong for larger parameters. The behavior-based grade requires as many variables and constants as the reference and searches at most 60 variable mappings, so an output that splits a variable or derives a declared constant fails even when its behaviors match. Its reverse check tests only the output's safety formula, so it accepts extra fairness conditions. \cib{2} \emph{Memorization} is the exposure effect of LoRA fine-tuning on each task three times with its test description. Contamination through public files exposes a model to the file alone, and our specification-only arm, which trains this way, stays too low on correctness to show an effect (see \autoref{tab:specexposure}). LoRA updates only small low-rank weights, so it may understate the memory that pretraining leaves. \cib{3} The \emph{cost of new names} is the decrease on unseen tasks. Random names also lengthen the token sequence, which the truncation analysis controls (see \autoref{app:truncation}), and CIR keeps the structure of a specification, so a model can still use a recalled structure.

\BfPara{External validity} All tasks come from one benchmark, \tla-Bench, and the 111 reference specifications are mostly single-module. The exposure experiment uses two open model families, and the cost of new names differs between them, 4.9 points for Qwen and 1.3 points for Gemma, so its size depends on the model and may differ for frontier models.

\BfPara{Statistical conclusion validity} The main threat is sample size, because public \tla specifications are few and \tla-Bench has only 146 model-checked system specifications. The behavior-based grade covers 90 tasks, the group with every required name stated has 33, and the pass rates of the exposure experiment lie between 7\% and 23\%, so a decrease of a few points falls inside the intervals, and the DiD on correctness includes 0 for both families. The likelihood measure does not depend on correct outputs and shows the exposure effect and its reduction under renaming clearly. We report several comparisons without a correction for multiple tests, and the bootstrap resamples tasks and misses the variation between training runs, since each exposure model is trained once with one split and one seed.

\section*{Ethical Considerations}

The specifications come from \tla-Bench, which collects them from 13 public repositories whose licenses permit redistribution with attribution~\cite{tlabench2026}. Each renamed copy is derived from one of these specifications, so the license of the original specification applies to it. The study involves no human participants and collects no personal data. We queried Opus 4.8 and GPT-5 through their commercial APIs under the providers' terms of use.

\section*{Acknowledgments}

We used a generative AI tool to assist with editing the text and with writing analysis code. The authors reviewed all text, code and results and take full responsibility for them.

\bibliography{references}

\clearpage
\appendix
\counterwithin{figure}{section}
\counterwithin{table}{section}
\startcontents[appendix]
\section*{Appendix Contents}
\printcontents[appendix]{}{1}{\setcounter{tocdepth}{1}}
\clearpage

\section{Prompt}
\label{app:prompt}

We query Opus 4.8 through the Anthropic Message Batches API and GPT-5 through the OpenAI Batch API, each with its default sampling settings. The frontier models receive the following instruction, with the description in place of \ttc{\{description\}}.

\begin{promptbox}
You are a TLA+ specification engineer. You will receive a natural language description of a system. Produce a complete, syntactically correct TLA+ specification that faithfully captures the described behavior. The output must be a valid TLA+ module that passes the SANY parser. Include the MODULE declaration, EXTENDS, CONSTANTS, VARIABLES, Init, Next and Spec. Include all safety invariants and liveness properties described and any fairness conditions mentioned. Output only the TLA+ specification.\\ 
~~ \\
System description: \{description\}
\end{promptbox}

\section{Properties That TLC Checks}
\label{app:properties}

TLC checks two kinds of properties, \cib{1} invariants, which must hold in every state the system reaches, and \cib{2} temporal properties, which must hold over whole behaviors. For the clock, an invariant called \ttc{TypeOK} could state that \ttc{hr} is always a number from 1 to 12. A temporal property can be a safety property, such as that a process never leaves a barrier early, or a liveness property, such as that the clock eventually shows 12. A clock that stops ticking may never reach 12, so a liveness property needs a fairness condition in the specification formula, which states that the clock keeps ticking~\cite{lamport2002specifying}.

\section{Identifier Kinds}
\label{app:idkinds}

\begin{table}[h]
\centering
\footnotesize
\setlength{\tabcolsep}{2.5pt}
\begin{tabular}{llrr}
\toprule
\tabhead{Kind} & \tabhead{Example} & \tabhead{Renamed} & \tabhead{Hidden} \\
\midrule
Domain words & \ttc{BarrierProperty} & 599 (34\%) & 148 (44\%) \\
Conventions & \ttc{TypeOK} & 415 (23\%) & 77 (23\%) \\
Abbreviations & \ttc{hr} & 774 (43\%) & 108 (32\%) \\
\midrule
Total & & 1,788 & 333 \\
\bottomrule
\end{tabular}
\caption{Identifier kinds in the 111 test tasks, over renamed and hidden required names.}
\label{tab:idkinds}
\end{table}

To see which identifiers are made of English words, we break each identifier into parts, so \ttc{FinishUpdate} becomes \emph{Finish} and \emph{Update} and \ttc{max\_val} becomes \emph{max} and \emph{val}. An identifier is a domain word when all its parts are English dictionary words.

\section{Renaming Example}
\label{app:hourclock}

\begin{figure*}[h]
\centering
\begin{minipage}[t]{0.48\textwidth}
\begin{codepanel}{gray}{black}{Original}
\small
\begin{alltt}
---- MODULE \idb{HourClock} ----
EXTENDS Naturals
VARIABLE \idb{hr}
\idb{HCini} == \idb{hr} \bs{}in (1 .. 12)
\idb{HCnxt} == \idb{hr}' = IF \idb{hr} # 12
              THEN \idb{hr} + 1 ELSE 1
\idb{HC} == \idb{HCini} /\bs [][\idb{HCnxt}]_\idb{hr}
==========================
\end{alltt}
\end{codepanel}
\end{minipage}\hfill
\begin{minipage}[t]{0.48\textwidth}
\begin{codepanel}{gray!20!white}{black}{Renamed}
\small
\begin{alltt}
---- MODULE \ido{Yodaczsbv} ----
EXTENDS Naturals
VARIABLE \ido{Tp}
\ido{Dylha} == \ido{Tp} \bs{}in (1 .. 12)
\ido{Azalu} == \ido{Tp}' = IF \ido{Tp} # 12
              THEN \ido{Tp} + 1 ELSE 1
\ido{Wc} == \ido{Dylha} /\bs [][\ido{Azalu}]_\ido{Tp}
==========================
\end{alltt}
\end{codepanel}
\end{minipage}
\Description{Two TLA+ modules side by side. The left is the HourClock specification. The right is
the same specification with every author-defined identifier renamed.}
\caption{The \ttc{HourClock} specification (left) and its renamed copy (right). Every
author-defined identifier (shaded) changes. Keywords and the standard module \ttc{Naturals} stay the same.
The renamed copy passes SANY and TLC with the renamed configuration.}
\label{fig:hourclock}
\end{figure*}

\section{Fine-Tuning Settings}
\label{app:finetune}

The two Qwen models use LoRA, which keeps all 14B base weights frozen and trains only small low-rank updates to them~\cite{hu2022lora} (rank 16, $\alpha=32$, dropout 0.05) on the query, key, value and output projections. Training runs for three epochs with batch size 1, the AdamW optimizer, learning rate $2\times10^{-4}$ with 100 warm-up steps and a linear decay, gradient clipping at 1.0, sequences truncated to 2,048 tokens, fp16 weights, gradient checkpointing and seed 0, so each test task in a model's half is seen three times. The loss covers only the specification tokens, and training runs on NVIDIA Quadro RTX GPUs with 48 GB of memory using Hugging Face Transformers and PEFT. The models are trained and tested with the prompt ``\ttc{Generate a TLA+ specification for the following system:}'', followed by the description and the line ``\ttc{TLA+ Specification:}''.

\section{Gemma 4 Exposure Experiment}
\label{app:gemma}

We fine-tune two copies of Gemma 4 31B~\cite{gemma4} on the same halves as the Qwen models, with the training settings of \autoref{app:finetune}, LoRA on the attention projections of the text layers and bfloat16 weights. Each model writes five samples per task and condition with temperature 0.7 and top-$p$ 0.95, and generation stops at the closing line of the module. \autoref{tab:gemma} gives the pass rates and effects.

\begin{table}[h]
\centering
\footnotesize
\setlength{\tabcolsep}{3pt}
\begin{tabular}{lrr}
\toprule
\tabhead{Measure} & \tabhead{Behavior grade} & \tabhead{Name grade} \\
\midrule
Tasks & 90 & 110 \\
Original, seen / unseen & 22.0 / 14.0 & 19.5 / 11.6 \\
Renamed, seen / unseen & 17.1 / 12.7 & 9.5 / 7.5 \\
Exposure effect & 8.0 [3.6, 12.9] & 7.8 [3.6, 12.4] \\
DiD & 3.6 [$-$0.9, 8.2] & 5.8 [1.5, 10.5] \\
Unseen cost & 1.3 [$-$2.0, 4.9] & 4.2 [0.4, 8.5] \\
\bottomrule
\end{tabular}
\caption{Gemma 4 31B under task-pair exposure, in percent and percentage points with 95\% CIs.}
\label{tab:gemma}
\end{table}

\section{Renaming Details}
\label{app:renaming}

We fix the random seed, so running the renaming again gives every identifier the same new name, and others can rebuild our exact test set. \tla keywords such as \ttc{VARIABLE}, \ttc{EXTENDS}, and \ttc{UNCHANGED} and standard module names such as \ttc{Naturals}, \ttc{Sequences}, and \ttc{FiniteSets} are excluded from renaming, since they belong to the language.

We keep the length because SANY reads the layout of a formula. A list of conditions joined by \ttc{/\bs} is grouped by the column in which each \ttc{/\bs} starts, so in \ttc{Init == /\bs{} x = 0} followed by \ttc{/\bs{} y = 0} on the next line, both \ttc{/\bs} must start in the same column. A longer or shorter name before the first \ttc{/\bs} would move it and change how SANY parses the formula. The renaming procedure also handles subscript forms that \tla derives from variables, such as \ttc{WF\_vars}.

In the reference specifications, renaming changes a median of 11 identifiers per specification, from 1 to 135. Only these names change, so a renamed copy still shares most of its text with the original specification, and the median word-by-word similarity between the two texts is 0.81, by Python's \ttc{difflib}.

\BfPara{Renamed descriptions} Blind replacement turns \textit{``the folklore one-step consensus algorithm [\ldots] as described by Dobre and Suri''} into \textit{``Zwk folklore one-step consensus algorithm [\ldots] as described Yf Dobre and Suri''} and it turns \textit{``Lamport's Bakery algorithm''} into \textit{``Lamport's Yodacz algorithm''}.

\section{Grading Details}
\label{app:grading}

\BfPara{Name-based grade} The instruction asks for operators called \ttc{Init}, \ttc{Next}, and \ttc{Spec}, and a renamed configuration names them by their new names, as in \ttc{INIT Octv}. The name-based grade therefore points the configuration's \ttc{INIT}, \ttc{NEXT} and \ttc{SPECIFICATION} entries at the output's \ttc{Init}, \ttc{Next} and \ttc{Spec} whenever the output defines them. An invariant or property entry keeps its name. If the output does not define that name and defines exactly one operator other than \ttc{Init}, \ttc{Next} and \ttc{Spec}, the entry names that operator. Constants keep their names.

\BfPara{Behavior-based grade} TLC checks these with $R$'s own definitions through an \ttc{INSTANCE} of $R$. The grader locates $G$'s specification formula as the operator the configuration names, an operator called \ttc{Spec} or any definition of the shape $I \land \Box[N]_v$. The configuration's constant values, state constraints, and symmetry sets apply to both checks.

\section{Control-Label Variables in the Behavior-Based Grade}
\label{app:labels}

Some variables have string values that act as control labels, as in the \ttc{pc} variable of a PlusCal translation. Their values depend on the label names, which the description does not fix. For a reference with such variables, the grader also tries 1-to-1 mappings between the string values of the two modules. If no mapping passes, it compares the two modules on their data variables. TLC writes the state graph of each module, the grader projects every state onto the data variables, and the two projected graphs must accept the same sequences of observable states. In one reference, the PlusCal translator names a step \ttc{"Lbl\_1"}. An output that names the same step \ttc{"start"} passes under the mapping from \ttc{"Lbl\_1"} to \ttc{"start"}.

\section{Validation of the Behavior-Based Grade}
\label{app:validation}

This experiment depends on the behavior-based grade, so we check that it accepts what it should and rejects what it should. We validate it on the 110 tasks that have a configuration. Of these, 8 stay in the test set for name-based grading, since they pass SANY and TLC after renaming and are valid tasks under benchmark practice, and we exclude them from behavior-based grading only. That grade needs \cib{1} TLC results that repeat across runs, which the specifications that use random values do not give, and \cib{2} system behavior to compare, which the test fixtures of the \tla tools do not describe. This leaves 102 tasks eligible for behavior-based grading. The grader accepts 90/102 references against themselves, and the other 12 use constructs that it does not translate, so the behavior-based grade covers 90/111 tasks (see \autoref{tab:counts}). The remaining task has no TLC configuration, so every output for it fails the name-based grade in every condition and does not change any decrease. For positive controls, we grade each reference against itself, a copy of each reference with its operators renamed to random names of the same width, and each renamed reference together with the renamed modules it imports. An operator-renamed copy counts only if it passes SANY and TLC, so operators that the configuration overrides by name or that an \ttc{INSTANCE} substitutes by name keep their names. For negative controls, we mutate each reference. A mutation changes a numeric literal by 1, flips a comparison, replaces a string value in the initial state, or deletes a conjunct of an action, a guard of an action, or a disjunct of the next-state relation. We keep a mutant only if TLC reports a different number of distinct reachable states for it, counted on the data variables for references with control-label variables. Every kept mutant thus behaves differently from its reference, and a correct grader must reject it. This selection makes rejection the easy direction, because TLC already exhibits the difference on the tested instance and the controls include no restructured but behaviorally equivalent specifications beyond renamings, so the false-negative rate on outputs that differ structurally from the reference is known only from the disagreements inspected in \autoref{sec:exp3-findings}.

The grader accepts 90/102 eligible references against themselves (see \autoref{tab:validation}). Of the 90 operator-renamed copies, 13 have no operator to rename, 3 do not parse and 18 fail TLC, so 56 are valid under that check. All 56 valid operator-renamed copies pass and all 90 renamed references pass together with the renamed modules they import. The grader rejects all 105 kept mutants, which come from the reference specifications of 54 tasks.

\begin{table}[h]
\centering
\small
\caption{Validation of the behavior-based grade on the 102 eligible tasks. A positive control should
pass and a negative control should fail.}
\label{tab:validation}
\setlength{\tabcolsep}{4pt}
\begin{tabular}{lrr}
\toprule
\tabhead{Control} & \tabhead{$n$} & \tabhead{Passed} \\
\midrule
Reference against itself & 102 & 90 \\
Operator-renamed reference & 56 & 56 \\
Renamed reference and modules & 90 & 90 \\
Mutant with other reachable states & 105 & 0 \\
\bottomrule
\end{tabular}
\end{table}

\section{Task Counts}
\label{app:counts}

\begin{table*}[t]
\centering
\small
\caption{Task counts. Indented rows are subsets of the row above, and the last two rows split the
test set. The last column gives what each set is used for.}
\label{tab:counts}
\begin{tabular}{lrl}
\toprule
\tabhead{Set} & \tabhead{$n$} & \tabhead{Used for} \\
\midrule
\tla-Bench system specifications & 146 & starting pool \\
\quad Pass SANY and TLC after CIR (test set) & 111 & Experiment 1 \\
\quad\quad With a TLC configuration & 110 & Experiment 2, name-based grading \\
\quad\quad\quad No test fixtures or random values & 102 & grader validation \\
\quad\quad\quad\quad Reference passes behavior-based grading & 90 & Experiments 2 and 3, behavior-based grading \\
\quad With a valid meaningful renaming & 77 & meaningful-name control \\
\midrule
With a hidden required name & 78 & hidden-name split (Experiment 1) \\
With every required name stated & 33 & hidden-name split (Experiment 1) \\
\bottomrule
\end{tabular}
\end{table*}

\section{File Continuation}
\label{app:continuation}

\BfPara{Setup} For each specification we show the model its first 40\% of lines (at least 8 and at most 60), tell it the repository and path the file comes from and ask it to continue the file exactly. The repository and path help a model that saw the file in training recall it~\cite{golchin2024timetravel}. We remove comments and separator lines from the model's output and from the rest of the real file, and we count the longest stretch of tokens that the two texts share word for word, so a long shared stretch suggests the model has seen the file. We run the same test on the renamed copy, which no model can have seen because it exists only in our study, so its shared stretches show how much matching comes from the common \tla text alone.

\BfPara{Findings} Opus 4.8 continues 46/111 specifications with a run of at least 50 tokens identical to the real file, and 25 renamed copies, which exist nowhere online, also produce such a run, since \tla modules share boilerplate. The 46 tasks whose file has a long run lose 8.7 points of correctness under renaming, and the other 65 lose 11.1 points (permutation $p=0.70$). The run on the original minus the run on the renamed copy has a Spearman correlation of 0.07 with the decrease (permutation $p=0.48$), and the 11 tasks above the 95th percentile of runs on renamed copies lose 0.0 points (permutation $p=0.25$). The continuation test therefore does not predict which tasks lose correctness. GPT-5 continues 11 specifications with a run of at least 50 tokens.

\section{Known Algorithm Names}
\label{app:algonames}

\BfPara{Setup} The renamed descriptions keep the names of published algorithms, protocols and classic problems, such as Paxos, the Bakery algorithm, Euclid's algorithm or the dining philosophers, and a model could use such a name to recall a known specification. We mark the tasks whose renamed description contains a name from a fixed list, which gives 37/111 tasks, and a broader list that adds problem words such as consensus, mutual exclusion and spanning tree gives 48/111. We compare the original correctness and the decrease after renaming between the marked tasks and the others, with a permutation test on the difference between the two mean decreases.

\BfPara{Findings} Tasks with a known name are harder. Under the behavior-based grade, Opus 4.8 is correct on 13.8\% of samples for the 32 covered tasks with a known name and on 52.1\% for the other 58. The decrease after renaming is 1.2 points (95\% CI $-5.6$ to 7.5) on tasks with a known name and 4.1 points (95\% CI $-0.7$ to 10.0) on the others (permutation $p=0.59$), and under the name-based grade it is 7.0 and 11.6 points ($p=0.45$). The broader list gives $p=0.54$ and $p=0.60$, and GPT-5 shows no difference under either grade or list ($p \geq 0.22$). A known name thus neither makes a task easier nor protects it from the decrease, so recall through such names does not explain our results. The marked tasks are also the more complex ones, so this comparison cannot rule out a small effect of the names.

\section{Truncation}
\label{app:truncation}

\BfPara{Setup} A random name such as \ttc{Bvwcdylhaazalut} splits into many short tokens, so a renamed specification has more tokens, 1,630 on average against 1,457 for the original. The fine-tuned models stop at 1,024 new tokens, so a renamed output is more likely to be cut off and fail, which could lower correctness after renaming without any role for memory. We count the outputs that reach the limit or lack the closing line of a module, and we repeat the analysis on the tasks with no such output in any condition.

\BfPara{Findings} Truncated outputs make up 16.9\% and 13.5\% of the original outputs of the models trained on half A and half B, and 18.7\% and 16.9\% of the renamed ones. Truncation is slightly more frequent in the renamed condition. On the 43/90 tasks with no truncated output in any condition, the renaming cost on unseen tasks is 9.8 points (95\% CI 3.7 to 16.3), so the renaming cost persists on tasks with no truncated outputs. The difference in differences on these tasks is 7.4 points (95\% CI 0.0 to 15.3), with an interval that reaches 0. The 43 tasks are not a random sample, because they are the shorter ones, and shorter specifications may be easier to memorize. The difference in differences is positive in every analysis, and its interval excludes 0 only under the name-based grade, so on correctness the evidence that renaming removes the gain from exposure is weak.

\section{Specification-Only Exposure}
\label{app:speconly}

\BfPara{Setup} In task-pair exposure, a seen specification appears in training next to its test description, so we also train a second pair of models with the same split, training settings, and 292 shared description-specification pairs. Each half of the test set enters training as the raw \ttc{.tla} file alone, with no description and no prompt. We call this a specification-only exposure. We test and grade these models in the same way. A model that learns from public repositories sees a specification file in the same way, so this arm measures the same three quantities under the kind of exposure that contamination usually produces.

\BfPara{Findings} With the original descriptions, the seen model is correct on 14.2\% of samples and the unseen model on 11.6\%, an exposure effect of 2.7 points whose interval includes 0. On unseen tasks, renaming costs 0.4 points (95\% CI $-3.6$ to 4.4) and the difference in differences is 2.7 points with an interval that includes 0. On correctness all three effects in this arm have intervals that include 0 (see \autoref{tab:specexposure}). On the likelihood of the reference, renaming removes part of the gain in this arm too, with a difference in differences of 0.08 (95\% CI 0.07 to 0.10). On the tokens outside the renamed identifiers it is 0.04 (95\% CI 0.03 to 0.05) and positive for 88/110 tasks. Correctness in this experiment stays between 11.6\% and 14.2\%, which leaves little room to measure a decrease. The seen model assigns the higher log-likelihood to the reference for 106/110 tasks.

\begin{table*}[t]
\centering
\small
\caption{Task-pair and specification-only exposure on the 90 tasks the behavior-based grade
covers. Task-pair models saw each specification with its test description, three times. Specification-only
models saw the raw \ttc{.tla} file, three times. Effects are in percentage points with 95\% bootstrap CIs,
except the log-likelihood row, which uses the mean token log-likelihood of the reference.
The last row counts the tasks for which the seen model assigns the higher likelihood to the
reference.}
\label{tab:specexposure}
\setlength{\tabcolsep}{5pt}
\begin{tabular}{L{6.0cm}L{3.1cm}L{3.1cm}}
\toprule
\tabhead{Measure} & \tabhead{Task-pair exposure} & \tabhead{Specification-only exposure} \\
\midrule
Exposure effect, original descriptions & 7.6 [3.8, 12.0] & 2.7 [$-$0.2, 5.8] \\
Exposure effect, renamed descriptions & 4.2 [1.1, 7.8] & 0.0 [$-$3.1, 3.3] \\
Difference in differences & 3.3 [$-$0.2, 7.3] & 2.7 [$-$1.8, 7.3] \\
Renaming cost, unseen tasks & 4.9 [2.0, 8.4] & 0.4 [$-$3.6, 4.4] \\
Log-likelihood, difference in differences & 0.16 [0.13, 0.19] & 0.08 [0.07, 0.10] \\
Log-likelihood, tokens outside renamed identifiers & 0.07 [0.06, 0.09] & 0.04 [0.03, 0.05] \\
Log-likelihood higher when seen (tasks) & 108/110 & 106/110 \\
\bottomrule
\end{tabular}
\end{table*}

\section{Disagreements Between the Two Grades}
\label{app:disagree}

The two grades disagree in both directions. Of the 450 Opus 4.8 samples per condition, 47 original and 71 renamed samples pass only the behavior-based grade, and 17 and 8 pass only the name-based grade. Behavior-only passes are mostly outputs that miss a hidden name. Reading the specifications behind the name-only passes shows that some describe different behavior. In one task the outputs fix a flag that the reference leaves free in the initial state, and in others an action differs. The name-based grade accepts them because TLC checks the outputs' own invariants. The other name-only passes come from limits of the behavior-based grade. These outputs \cib{1} encode a value differently, such as \ttc{TRUE} and \ttc{FALSE} for the 0 and 1 of the reference, \cib{2} declare an extra constant or derive one that the reference declares or \cib{3} write the specification formula in a form the grader does not parse. Most GPT-5 name-only passes come from one task with the boolean encoding. If every name-only pass counts as correct, which also credits the outputs with different behavior, the decrease for Opus 4.8 is 5.1 points (95\% CI 0.4 to 10.2), half of the 10.4-point decrease under the name-based grade, and for GPT-5 it is 0.4 points (95\% CI $-3.1$ to 4.0). The 90\% interval of the Opus 4.8 decrease runs from $-0.2$ to 6.7 points. Its upper end exceeds the 5-point margin, so the data do not rule out an effect on Opus 4.8.

\section{Grading in Public \tla Benchmarks}
\label{app:benchmarks}

\begin{table}[h]
\centering
\footnotesize
\setlength{\tabcolsep}{3pt}
\begin{tabular}{L{2.7cm}L{3.2cm}L{1.0cm}}
\toprule
\tabhead{Benchmark and task} & \tabhead{Check} & \tabhead{Hidden name} \\
\midrule
\tla-Bench~\cite{tlabench2026}, description to specification & SANY and TLC with the reference configuration & 78/111 \\
SysMoBench~\cite{sysmobench2026}, code to system model & SANY, TLC, trace validation and invariant templates, with names mapped by an LLM & no \\
TLAPS-Bench~\cite{tlapsbench2026}, theorem to proof & TLAPS on the stated theorem & no \\
TLAi+Bench~\cite{tlaibench2025}, puzzle to specification & TLC counterexample and a refinement mapping to the gold specification & no \\
\bottomrule
\end{tabular}
\caption{Whether four \tla benchmarks need a name that the task does not state.}
\label{tab:benchmarks}
\end{table}

\section{Meaningful Names}
\label{app:meaningful}

\BfPara{Setup} Random names might lower correctness because they remove the information that readable names carry, so we built a third condition to test this. Claude Opus 4.8 proposes a new meaningful name for each identifier and we apply and re-verify the renaming in the same way as the random one. We built these maps for the first 100 tasks of the test set, before it grew to 111, and 77/100 pass SANY and TLC after renaming, so the meaningful condition has 77.

\BfPara{Name-based grade} On the 77 tasks with a valid meaningful renaming, Opus 4.8 is correct on 30.1\% of samples with the original descriptions, on 19.7\% with random names, and on 20.0\% with meaningful names. The difference between random and meaningful names is $-0.3$ points (95\% CI $-2.3$ to $1.6$). Its 90\% interval runs from $-2.1$ to $1.3$ and lies inside the margin of 5 points, so under the name-based grade the two renamings are equivalent, since both replace the same hidden names. For GPT-5, random and meaningful names give the same rate, 7.3\%, on the 77 tasks with both renamings.

\BfPara{Behavior-based grade} For Opus 4.8, the meaningful renaming is valid on 61 of these tasks. On them, the behavior-based rate with meaningful names is 37.7\%, 1.3 points (95\% CI $-3.6$ to 6.2) below the rate of 39.0\% with random names. The upper end exceeds the margin of 5 points, so under the behavior-based grade the comparison of the two renamings is inconclusive. On the 61 tasks with a meaningful renaming, GPT-5 is correct on 10.8\% with random names and 11.1\% with meaningful names.

\section{Additional Results of Experiment 1}
\label{app:exp1}

\begin{figure*}[h]
\centering
\includegraphics[width=\textwidth]{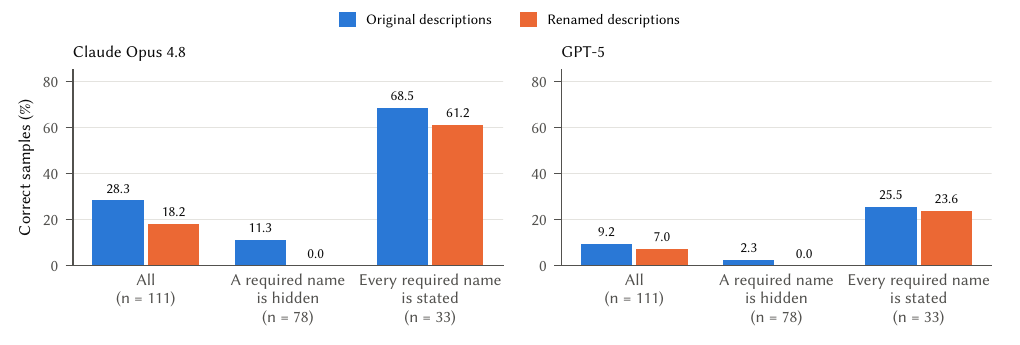}
\Description{Grouped bar charts, one per model, of correctness with original and renamed descriptions for all 111 tasks, the 78 with a hidden required name and the 33 whose descriptions state every required name.}
\caption{Frontier models under the name-based grade, five samples per task, split by whether a required name is hidden.}
\label{fig:hidden}
\end{figure*}

The decrease of Opus 4.8 also appears when each of the five sample positions is analyzed separately. On the 100 tasks that \tla-Bench reports, the first original sample passes for 27\% with Opus 4.8 and 10\% with GPT-5. \tla-Bench reports 16\% for Opus 4.5 and 4\% for GPT-5 on the same tasks, with its own prompt and generation settings, so the rates are not directly comparable. BLEU-4 and ROUGE-L decrease between 2.6 and 5.4 points under renaming (see \autoref{tab:similarity}), so outputs stay close to the reference text when the names change.

\section{Additional Results of Experiment 2}
\label{app:exp2}

On the tokens of the renamed identifiers alone, the exposure effect on the likelihood of the Qwen models is 0.00 (95\% CI $-0.04$ to 0.03).

\begin{figure*}[h]
\centering
\includegraphics[width=\textwidth]{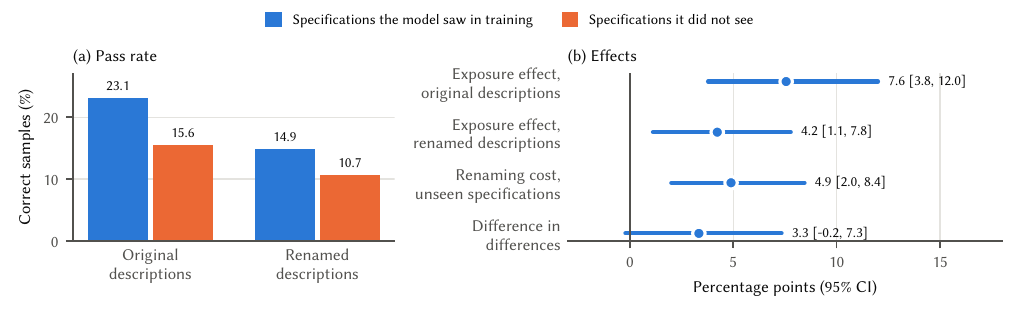}
\Description{Left, grouped bars of pass rates for seen and unseen tasks with original and
renamed descriptions. Right, point estimates with 95 percent intervals for the exposure effects, the
renaming cost on unseen tasks and the difference in differences.}
\caption{Randomized exposure on the 90 tasks the behavior-based grade covers, with two Qwen2.5-Coder-14B models trained on disjoint halves. (a) Behavior-based pass rate. (b) Effects with 95\% CIs over tasks.}
\label{fig:exposure}
\end{figure*}

\section{Additional Results of Experiment 3}
\label{app:exp3}

\begin{figure*}[h]
\centering
\includegraphics[width=\textwidth]{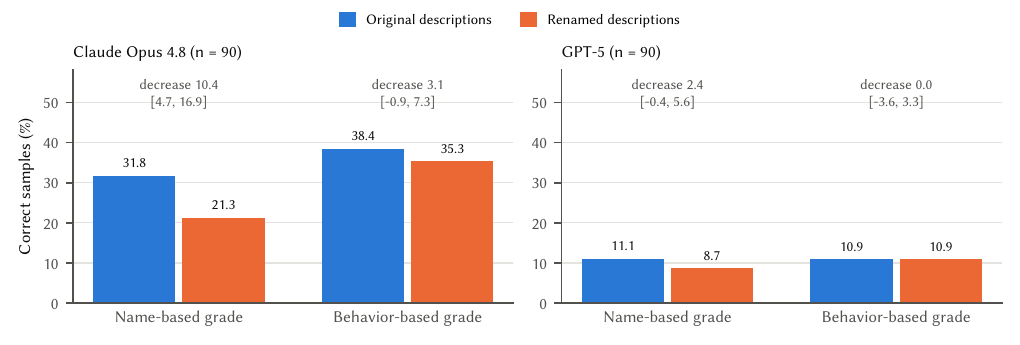}
\Description{Grouped bar charts, one per model, of correctness with original and renamed descriptions,
under the name-based grade and the behavior-based grade.}
\caption{Frontier models on the 90 tasks the behavior-based grade covers, graded by name and by behavior. Each decrease is original minus renamed with a 95\% CI.}
\label{fig:graders}
\end{figure*}

On the 90 tasks that both grades cover, renaming makes 15 tasks of Opus 4.8 fail and two pass under the name-based grade, and nine fail and three pass under the behavior-based grade. For Opus 4.8, on the 64 of these tasks with a hidden required name, the name-based grade gives 12.5\% and 0.0\%, and the behavior-based grade gives 22.8\% and 20.3\%, a difference of 2.5 points (95\% CI $-1.9$ to 7.2). On the 26 whose descriptions state every required name, the behavior-based grade gives 76.9\% and 72.3\% (difference 4.6 points, 95\% CI $-4.6$ to 13.8).

\section{Text Similarity}
\label{app:similarity}

BLEU-4 is the overlap of 1 to 4 token sequences between the output and the reference~\cite{papineni2002bleu}, and ROUGE-L is the longest common token subsequence of the two texts~\cite{lin2004rouge}. Both are computed against the reference of the same condition without comments.

\begin{table}[h]
\centering
\footnotesize
\setlength{\tabcolsep}{4pt}
\begin{tabular}{lrrr}
\toprule
\tabhead{Measure} & \tabhead{Original} & \tabhead{Random} & \tabhead{Meaningful} \\
\midrule
\multicolumn{4}{l}{\emph{Claude Opus 4.8}} \\
BLEU-4 & 47.3 & 41.9 & 44.3 \\
ROUGE-L & 60.0 & 56.9 & 58.6 \\
\midrule
\multicolumn{4}{l}{\emph{GPT-5}} \\
BLEU-4 & 36.1 & 32.1 & 33.9 \\
ROUGE-L & 51.0 & 48.4 & 49.9 \\
\bottomrule
\end{tabular}
\caption{Text similarity of frontier outputs to the reference of their condition, after comments are removed, in percent, averaged over samples and then tasks.}
\label{tab:similarity}
\end{table}

\end{document}